\documentclass[letterpaper, 10 pt, conference]{ieeeconf}  

\IEEEoverridecommandlockouts                              

\usepackage{amsmath,amssymb,amsfonts}
\usepackage{algorithm}
\usepackage{algorithmic}
\usepackage{graphicx}
\usepackage{textcomp}
\usepackage{tabularx}
\usepackage{booktabs}
\usepackage{makecell}
\usepackage{caption}
\usepackage{xcolor}
\usepackage[colorlinks=true,urlcolor=blue,breaklinks=true]{hyperref}

\definecolor{Pink}{HTML}{F5E6EF}
\definecolor{Red}{HTML}{FEE3CE}
\definecolor{Blue}{HTML}{EAEDF2}
\definecolor{Green}{HTML}{CFEBDF}
\definecolor{Yellow}{HTML}{E8AE45}
\definecolor{Grey}{HTML}{BFBFBF}
\definecolor{Purple}{HTML}{C0AAF0}

\title{\LARGE \bf
KiRAS: Keyframe Guided Self-Imitation for Robust and Adaptive Skill Learning in Quadruped Robots
}

\author{Xiaoyi Wei$^{1}$, Peng Zhai$^{1,*}$, Jiaxin Tu$^{1}$, Yueqi Zhang$^{1}$, \\
Yuqi Li$^{1}$, Zonghao Zhang$^{1}$, Hu Zhou$^{2}$, and Lihua Zhang$^{1,*}$\\
Project Website: \protect\href{https://lainweigh.github.io/kiras}{\texttt{https://lainweigh.github.io/kiras}
\vspace{-0.3cm}
}
\thanks{$^{1}$College of Intelligent Robotics and Advanced Manufacturing, Fudan University, China. (email:
        {\tt\small weixy23@m.fudan.edu.cn; \{pzhai, lihuazhang\}@fudan.edu.cn.})}%
\thanks{$^{2}$Power China Huadong Engineering Corporation Limited, China.}%
\thanks{* Corresponding Author.}
}

\begin{document}
\maketitle
\thispagestyle{empty}
\pagestyle{empty}

\begin{abstract}

With advances in reinforcement learning and imitation learning, quadruped robots can acquire diverse skills within a single policy by imitating multiple skill-specific datasets. However, the lack of datasets on complex terrains limits the ability of such multi-skill policies to generalize effectively in unstructured environments. Inspired by animation, we adopt keyframes as minimal and universal skill representations, relaxing dataset constraints and enabling the integration of terrain adaptability with skill diversity.
We propose Keyframe Guided Self-Imitation for Robust and Adaptive Skill Learning (KiRAS), an end-to-end framework for acquiring and transitioning between diverse skill primitives on complex terrains. KiRAS first learns diverse skills on flat terrain through keyframe-guided self-imitation, eliminating the need for expert datasets; then continues training the same policy network on rough terrains to enhance robustness. To eliminate catastrophic forgetting, a proficiency-based Skill Initialization Technique is introduced. Experiments on Solo-8 and Unitree Go1 robots show that KiRAS enables robust skill acquisition and smooth transitions across challenging terrains. This framework demonstrates its potential as a lightweight platform for multi-skill generation and dataset collection. It further enables flexible skill transitions that enhance locomotion on challenging terrains.

\end{abstract}

\section{INTRODUCTION}

\begin{figure}[t]
\vspace{0.2cm}
\centerline{\includegraphics[width=\columnwidth]{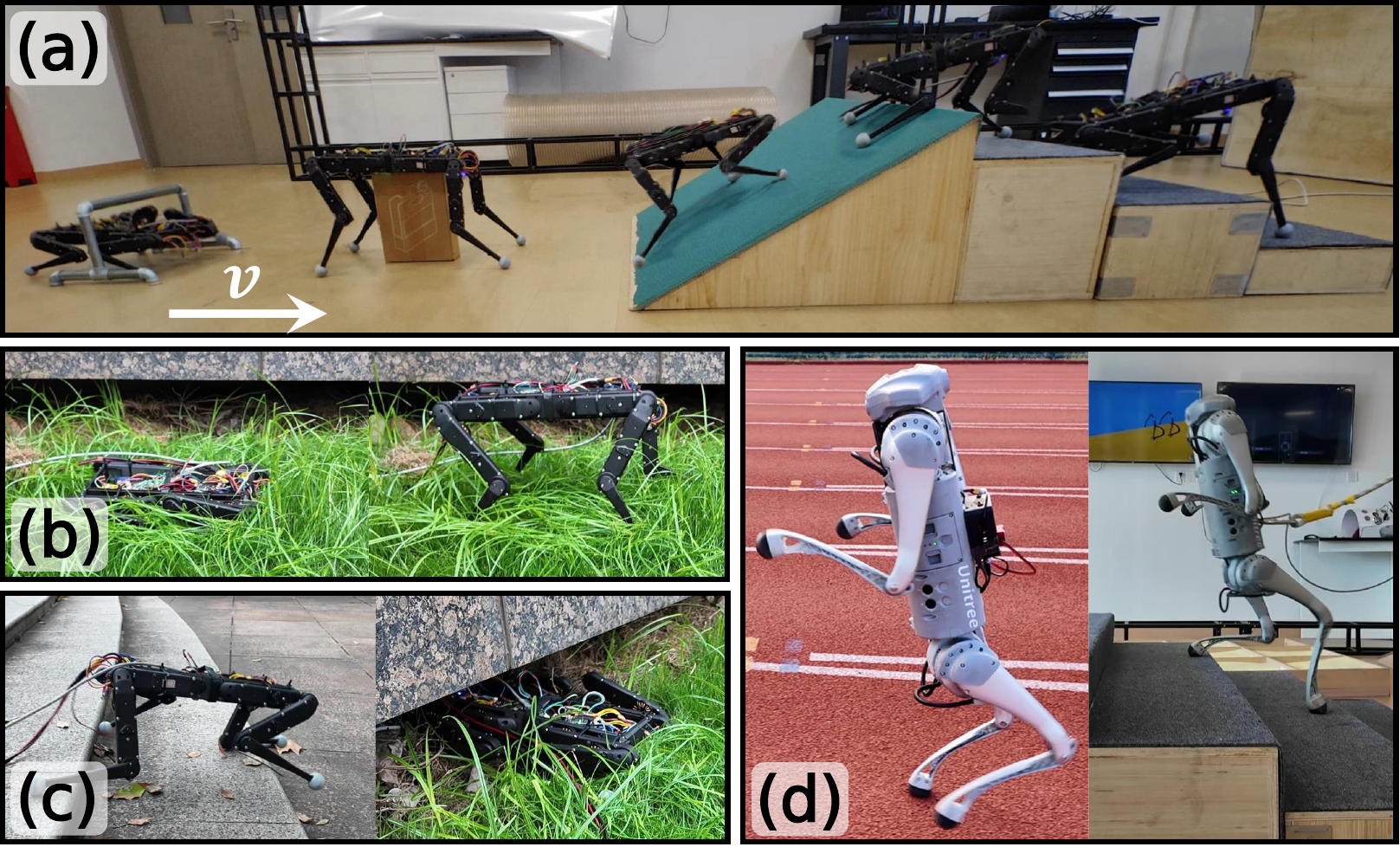}}
\captionsetup{font=footnotesize}
\caption{Deployment of KiRAS on Solo-8 and Unitree Go1 robots across diverse environments. (a) Solo-8 robot traverses various obstacles by flexibly switching learned skills. (b) Solo-8 robot exploits crawl for concealment and stilt for expanded field of view. (c) In outdoor unstructured environments, Solo-8 robot demonstrates strong robustness. (d) Unitree Go1 robot acquires bipedal skills to climb a 10\,cm step.
}
\vspace{-0.4cm}
\label{performance}
\end{figure}

Recent advances in deep reinforcement learning (DRL) have led to breakthroughs in quadruped locomotion, enabling quadruped robots to traverse unstructured terrains, such as stairs and obstacles, using proprioception~\cite{10.1126/scirobotics.abk2822, nahrendra2023dreamwaq}. However, these policies often learn a single terrain-specific solution, overlooking alternative behaviors required by different tasks. This limits their ability to meet diverse demands, such as crawling for concealment during wildlife photography (Fig.~\ref{performance}(b)) or adopting a bipedal stance for inspection tasks (Fig.~\ref{performance}(d))~\cite{niu2025human2locoman}. This gap underscores the need to integrate \textbf{terrain adaptability} with \textbf{skill diversity}.

Current approaches to diverse skill learning mainly rely on reward engineering~\cite{margolis2023walk} or dataset-driven imitation learning, such as Adversarial Motion Prior (AMP)~\cite{Peng_2021}. Both achieve reliable execution and stable switching of multiple skills on flat terrain, but extending them to complex terrains remains challenging. In unstructured settings, reward engineering must balance skill acquisition with terrain adaptation, greatly increasing design complexity. Imitation-based methods reduce this burden using expert datasets, yet these datasets strictly constrain \textbf{skill representation} and are mostly collected on flat terrain, rarely include rough-terrain scenarios. Consequently, directly applying flat-terrain datasets to rough terrains often leads to overfitting, as terrain-specific motion patterns are poorly captured, and desired motions cannot be reliably reproduced.
Although recent studies explore skill-conditioned traversal~\cite{han2024lifelike,zhang2025motion,huang2025moe}, they still face complex reward design and poor scalability. Thus, achieving terrain adaptability through motion imitation in challenging environments remains an open problem.

Given the scarcity of datasets on complex terrains, a critical question arises: is there a compact, general skill representation that can describe skills with minimal resources, thereby weakening the strong motion constraints imposed by full expert datasets? \textbf{Keyframe} from character animation offers one such idea: animators specify a few pivotal poses, and the system interpolates transitions between them, eliminating the need to animate every frame manually~\cite{sturman1984interactive}. The same idea naturally extends beyond animation: when acquiring new skills, humans also rely on a small set of decisive movements from demonstrations (videos, texts, or mentors), iteratively imitating, refining the successful instances, and generalizing to new situations~\cite{ijspeert2013dynamical}. Motivated by the shared principle, we treat keyframes as high-level objectives and introduce self-imitation learning (SIL,~\cite{oh2018selfimitationlearning}) as the underlying framework. During training, SIL selects the robot’s high‑return trajectories that meet keyframe targets and reinforces them, allowing natural and generalizable motion policies without constraints from fixed demonstrations or large datasets. This enables the robot to better adapt to diverse environmental dynamics and offers a promising path toward unifying terrain adaptability and skill diversity in legged locomotion.

We present \textbf{KiRAS} (Keyframe Guided Self-Imitation for Robust and Adaptive Skill Learning), an end-to-end framework for quadruped robots to traverse terrains with diverse skill primitives. Guided by high-level keyframes, KiRAS first acquires diverse skills and transitions via SIL on flat terrain, and then transfers the policy to rough terrains through same policy network for robustness. We further introduce a proficiency-based Skill Initialization Technique that prioritizes under-trained skills to mitigate catastrophic forgetting. Experiments on Solo-8~\cite{Grimminger_2020} and Unitree Go1 robots (Fig.~\ref{performance}) demonstrate the method’s ability to perform multi-skill terrain traversal and generalization across platforms.

The main contributions of this work are as follows:

\begin{itemize}
    \item A keyframe-guided pipeline that extracts high-quality trajectories for SIL to acquire skill primitives, and subsequently trains the same policy network for complex terrains using an Environmental Context Estimator (ECE) conditioned on proprioception.
    \item A proficiency-driven Skill Initialization Technique that adaptively oversamples under-mastered skills while preserving already acquired behaviors to balance learning.
    \item Comprehensive evaluations on Solo-8 and Unitree Go1 robots in both simulation and hardware, demonstrating state-of-the-art performance and flexible, robust multi-skill terrain traversal.
\end{itemize}

\section{RELATED WORK}

\subsection{Robust DRL Controller without External Sensors}
Since the introduction of Rapid Motor Adaptation~\cite{kumar2021rma}, which demonstrated that DRL policies outperform traditional controllers in robustness for quadrupedal locomotion, increasing attention has been devoted to enabling proprioceptive control and environmental awareness without external sensors. Ji~et~al.~\cite{ji2022concurrent} propose a state estimation network to infer linear velocity, foot height, and contact probability directly from onboard observations, reducing reliance on external inputs. Building on this, Nahrendra~et~al.~\cite{nahrendra2023dreamwaq} introduce DreamWaQ, enhancing estimation accuracy with a context-aided estimator. Long~et~al.~\cite{long2023hybrid} develop a model combining regression and contrastive learning to predict velocity and latent states from proprioceptive history. These proprioception-based modules enable the robot to estimate physical quantities not accurately observable through an IMU and motors, improving state awareness, terrain adaptability, and controller robustness.

However, most of these approaches adopt an all-in-one policy that learns a single fixed approach for each terrain via environmental adaptation. While this improves robustness to some extent, it limits the ability to select and switch skills according to task requirements. Consequently, robots rely on a single behavioral pattern per terrain, restricting flexibility and performance in multi-task scenarios, especially when skill switching is critical.

\subsection{Multi-skill Learning for Legged Robots}
Multi-skill learning aims to develop a unified policy network that can acquire and manage multiple skills. In legged robots, Generative Adversarial Imitation Learning (GAIL)~\cite{ho2016generative} is a mainstream approach that reproduces expert state-action distributions through adversarial training, eliminating the need for explicit reward design. AMP~\cite{Peng_2021} extends this idea by enabling skill learning from expert datasets. Building on this foundation, Li et al. propose a series of methods: WASABI~\cite{li2023learning} infers rewards from rough partial demonstrations; Cassi~\cite{10160421} integrates GAIL with unsupervised skill discovery to isolate individual skills from unlabelled data, while FLD~\cite{li2024fld} constructs a latent space to improve interpolation and generalization from sparse trajectories. Moreover, Vollenweider et al.~\cite{vollenweider2022advancedskillsmultipleadversarial} employ parallel discriminators for each skill to enable skill transition. Zargarbashi et al.~\cite{zargarbashi2024robotkeyframing} propose a keyframe-guided approach for natural gait generation with temporal keyframe constraints, though it still relies on AMP to compute style rewards. 

However, most methods focus on flat terrain and overlook the advantages of diverse skills in complex environments, leaving skill transition for high-level tasks underexplored. While recent studies have begun exploring this direction, they still face limitations. Margolis et al.~\cite{margolis2023walk} introduce a multi-skill policy modulated by behavior parameters, but it requires carefully crafted rewards and lacks terrain adaptability. Han et al.~\cite{han2024lifelike} and Zhang et al.~\cite{zhang2025motion} train multi-gait base networks on large-scale datasets and use residual networks for high-level control. While effective in complex terrains, their multi-stage distillation increases architectural complexity and reduces policy scalability. Huang et al.~\cite{huang2025moe} use a Mixture-of-Experts framework for quadrupedal–bipedal transitions, but skill and task diversity remain limited. In contrast, KiRAS acquires diverse skills from keyframes without relying on high-quality expert datasets or elaborate reward engineering. It employs an end-to-end training pipeline for a single policy network, balancing performance with architectural simplicity and avoiding the information loss typically introduced by distillation.

\begin{figure*}[htbp]
\vspace{0.2cm}
\centerline{\includegraphics[width=\textwidth]{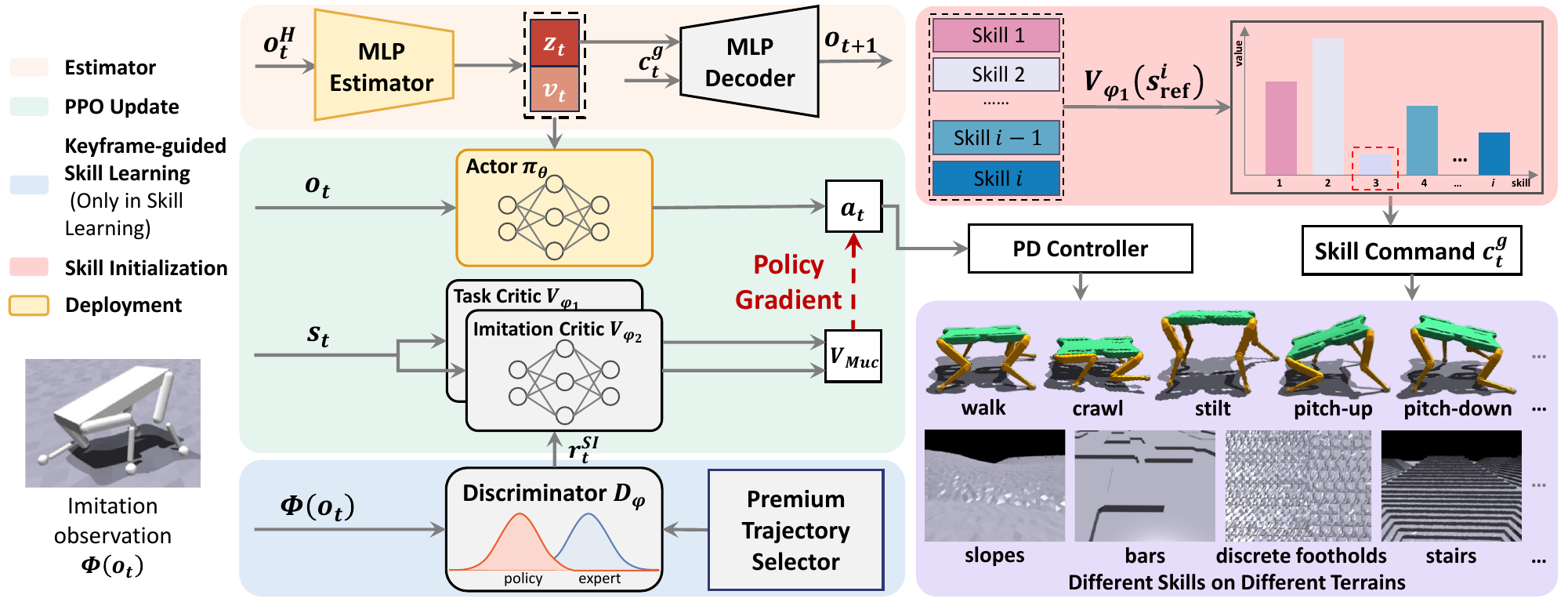}}
\captionsetup{font=footnotesize}
\caption{Training pipeline. KiRAS consists of an end-to-end adaptive multi-skill learning framework and a skill initialization module. The former includes a keyframe-guided skill learning module (used only in skill learning stage), a proprioception estimator, and a PPO actor-critic architecture. The policy outputs joint position targets, which are converted to torques via a PD controller. The Skill Initialization Technique is employed at environment reset to prevent overfitting to simpler skills. The \textcolor{Purple}{purple box} illustrates different skills being trained on different terrains. During deployment, only the networks in the \textcolor{Yellow}{yellow boxes} are used, and skill switching is controlled via joystick input.
}
\vspace{-0.6cm}
\label{method}
\end{figure*}

\section{METHOD}
In this section, we introduce KiRAS framework, as illustrated in Fig.~\ref{method}. We first present the training pipeline, which performs \textbf{skill learning} followed by \textbf{terrain finetuning}. Then, we describe the Skill Initialization Technique. Finally, we provide some key training details.

\subsection{End-to-end Adaptive Multi-skill Learning Framework}
KiRAS enables quadruped robots to perform various skills across diverse terrains, with smooth transitions between them. During \textbf{skill learning}, KiRAS learns these skills on flat terrain using keyframe-based SIL, where each skill is represented by \textit{one} single keyframe and encoded into the observation space via a one-hot vector $c_t^g$. While efficient, this minimalist representation limits adaptability: joint positions in a keyframe remain fixed, failing to capture differences in execution between, for example, walking on flat ground and ascending stairs. To overcome this limitation, the subsequent \textbf{terrain finetuning} disables SIL module and continues training the same policy network to acquire robustness and flexibility for terrain traversal. Unlike conventional two-stage frameworks that rely on distillation, KiRAS achieves joint learning of skills and terrain adaptability through the keyframe mechanism, mitigating inefficiency and information loss observed in traditional pipelines.

\subsubsection{Keyframe-guided Skill Learning}
\label{keyframe}

In skill learning, we use keyframes as inputs for SIL to facilitate adversarial skill acquisition. Our approach builds on the WASABI~\cite{li2023learning} framework. Keyframes can be seen as special cases of partial demonstration containing a single timestep. However, only repeating this static frame to form transitions can cause overfitting to the posture and lead to temporally incoherent actions. To solve this, we use premium trajectories collected through exploration as demonstrations for SIL.

\begin{figure}[t]
\vspace{0.2cm}
\centerline{\includegraphics[width=\columnwidth]{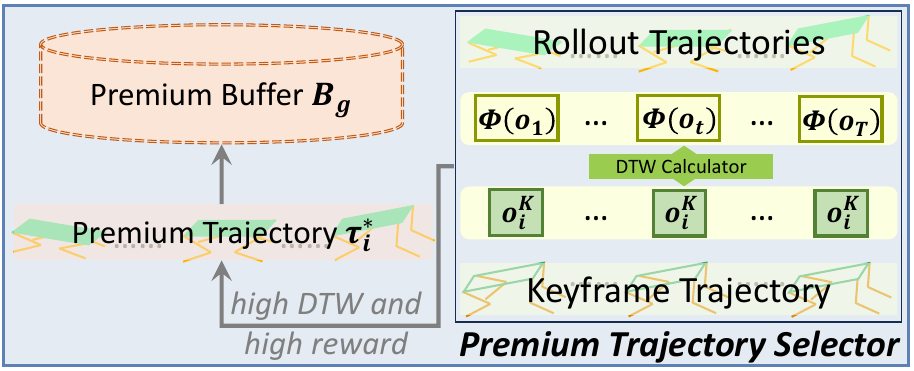}}
\captionsetup{font=footnotesize}
\caption{Details of Premium Trajectory Selector.}
\vspace{-0.8cm}
\label{premium-selector}
\end{figure}

Fig.~\ref{premium-selector} illustrates the Premium Trajectory Selector. For each skill $i$, we maintain a premium buffer $B_g = \{\tau_i^*\}$, where $\tau_i^*$ denotes premium trajectories. Premium buffer is initialized with trajectories generated by repeating the keyframe $K$ of skill $i$ for $T$ timesteps, referred to as \emph{keyframe trajectories}: $\{o_i^K\}^T$, where $o_i^K$ contains partial observations such as the root state and joint positions. Inspired by~\cite{tu2025continuous}, we select a \textit{premium trajectory} $\tau_\pi = (\Phi(o_1), \Phi(o_2), ..., \Phi(o_T))$ for skill $i$ based on two criteria: (\emph{i}) the combined reward $r_t^c$ accumulated over the rollout (detailed in Section.~\ref{reward}), measuring performance and alignment with desired behaviors, and (\emph{ii}) the Dynamic Time Warping (DTW) distance to the keyframe trajectory, which measures structural similarity in the imitation space. Here, $\Phi$ denotes an operator that extracts partial information from the full observations into the imitation space, matching the observation modality of the keyframe. The combined evaluation is defined as:
\begin{equation}
    J(\tau_\pi) = \sum_{t=0}^T r_t^c +  d^{\text{DTW}}\left(\tau_\pi, \{o_i^K\}^T\right). 
\end{equation}
After each episode, if $J(\tau_\pi)$ exceeds the current maximum score $\epsilon_i$ of the premium trajectory for skill $i$ in the premium buffer, the trajectory is added to $B_g$; otherwise it is discarded. Keyframe trajectories are consistently retained in the premium buffer as references.

We then train a discriminator $D_\psi$ to distinguish state transitions $(\Phi(o_{t-1}), \Phi(o_t))$ from the premium buffer distribution $B_g$ or the policy distribution $d_\pi$. Following the same formulation as in WASABI, SIL reward $r_t^{SI}$ encourages the policy to generate transitions resembling high-return trajectories, guiding self-imitation and effective skill learning.

\subsubsection{Rewards and Multi-critic Architecture}
\label{reward}
In KiRAS, the total reward comprises 4 components: SIL reward $r_t^{{SI}}$ for learning skill primitives, the termination penalty $r_t^T$, the regularization reward $r_t^R$ for stable real‑world deployment, and a novel residual penalty $r_t^{{res}}$ that constrains joint outputs. More details about $r_t^T$ and $r_t^R$ can be found in~\cite{li2023learning}, and the residual term $r_t^{{res}}$ is defined as:
\begin{equation}
r_t^{{res}} =
\begin{cases}
0, & 0 \le t < T_1,\\
- \| a - a_{\mathrm{flat}}\|, & T_1 \le t < T_2,
\end{cases}
\end{equation}
where $T_1$ and $T_2$ denote the timesteps of skill learning and terrain finetuning, respectively, and $a_{\mathrm{flat}}$ is the action produced in skill learning under identical observations. The reward aligns actions in terrain finetuning with those from skill learning to prevent detrimental deviations.

During terrain finetuning, all terms except $r_t^{{SI}}$ remain active to guide robust terrain adaptation. Relying on a single critic to estimate a composite value function can overload by heterogeneous reward signals, causing high variance in value estimates, and unstable policy gradients. Moreover, residual gradients from $r_t^{{SI}}$ can persist and disrupt terrain learning.

To address these issues, we adopt a multi-critic (MuC) architecture inspired by~\cite{zargarbashi2024robotkeyframing}, where each critic $V_{\phi_i}$ is responsible for estimating a specific subset of rewards. Specifically, the \textbf{Task Critic} $V_{\phi_1}$ estimates the cumulative return from the combined reward $r_t^c = \omega^T r_t^T + \omega^R r_t^R + \omega^{{res}} r_t^{{res}}$, while the \textbf{Imitation Critic} $V_{\phi_2}$ estimates the cumulative return from the imitation reward $\omega^{{SI}} r_t^{{SI}}$. Each critic is trained independently to specialize in its corresponding reward signal. At each update, we compute normalized advantages $\hat A_i$ with mean $\mu_i$ and standard deviation $\sigma_i$, and combine them as $\hat{A}_{{MuC}}= \sum_{i=1}^2 \omega_i \,\frac{\hat{A}_i - \mu_i}{\sigma_i}\,.$
However, unlike prior work, we dynamically adjust the mixing weights $\omega_1,\omega_2$ over training iterations $t$ with hyperparameter $\sigma$:
\begin{equation}
\begin{aligned}
\omega_1 &=
\begin{cases}
\frac{\sigma}{T_1}t + (1-\sigma)t, & 0 \le t < T_1,\\
1, & T_1 \le t < T_2,
\end{cases} \\
\omega_2 &= 1 - \omega_1 ,
\end{aligned}
\end{equation}
where $\omega_1$ and $\omega_2$ denote the task advantage and imitation advantage weight. This linear schedule shifts emphasis from skill imitation to terrain adaptation as training progresses.

\subsubsection{Environment Context Estimator}
To enhance terrain robustness and adaptability, we introduce ECE that enriches the policy network with proprioceptive context. ECE consumes the past $H=5$ steps of proprioceptive observations $o_t^H$ to predict the body velocity $v_t$ and produce a latent vector $z_t$, which is used to reconstruct the next observation $o_{t+1}$. Previous studies have demonstrated that accurate velocity estimation and implicit environmental representation are critical for locomotion~\cite{nahrendra2023dreamwaq, long2023hybrid}. However, different skills correspond to distinct postures and contact dynamics, and conditioning on the skill command  $c_t^g$ helps the estimator capture these variations across skills.

To address this, we employ a Conditional Variational Autoencoder (CVAE) that incorporates the skill command as a context prior. The observation reconstruction is defined as $\hat{o}_{t+1} = \mathrm{Decoder}(z_ t, c_t^g)$, and the velocity prediction is supervised using MSE. The combined loss is
\begin{align}
    \mathcal{L}_{\mathrm{ECE}} = &\; MSE(v_t, \hat{v}_t) + MSE(o_{t+1}, \hat{o}_{t+1}) \notag \\
    &+ \beta\,D_{\mathrm{KL}}\big(q(z_t\,|\,o_t^H, c_t^g)\,\|\,p(z_t\,|\,c_t^g)\big)\,,
\end{align}
where $\hat v_t$ and $\hat o_{t+1}$ are the predicted velocity and observation, respectively. $q(z_t\,|\,o_t^H, c_t^g)$ is the posterior distribution and $p(z_t\,|\,c_t^g)$ is a skill‑conditioned Gaussian prior.

Since skill learning is performed on flat terrain and terrain finetuning on complex terrain, early ECE convergence on flat terrain may hinder the modeling of uneven environments. Thus, we adopt adaptive sampling (AdaBoot), which conditionally updates the estimator based on reward ratio~\cite{nahrendra2023dreamwaq}, improving generalization to complex terrains.

\subsection{Skill Initialization Technique}
\label{skill_initialization}
One of the key challenges in multi-skill learning is exploration bias, where the policy tends to overfit to easier sub-skills while neglecting those that are more difficult to acquire~\cite{riemer2018learning}. To counteract this bias, we introduce the Skill Initialization Technique that probabilistically favors undertrained skills. Taking advantage of MuC framework, we leverage Task Critic to evaluate the performance of each skill and dynamically adjust their sampling probabilities based on their task values, which enables balanced training across all skills within the policy network, mitigating catastrophic forgetting and promoting comprehensive skill development.

Concretely, at the start of an episode, we associate each skill $i$ with a reference state $s_i^{\text{ref}}$, from which we compute its task value $V_{\phi_1}(s_i^{\text{ref}})$ using the Task Critic. In practice, we set $s_i^{\text{ref}}$ to the skill’s corresponding keyframe $s_i^K$. The probability of selecting skill $i$ at the beginning of an episode is:
\begin{equation}
p(\text{Skill}=i)=\frac{1}{N-1} \left( 1-\frac{V_{\phi_1}(s_i^{\text{ref}})}{\sum_{m=1}^N V_{\phi_1}(s_m^{\text{ref}})}\right)\,,
\end{equation}
where $N$ denotes the total number of skills. Skills with lower task values are more likely to be sampled, encouraging focused improvement. Using relative performance, the policy can adaptively shift attention toward weaker skills.

To ensure all skills remain represented during training, we perform a coverage check. If any skill has not been sampled recently, we substitute one of the most frequently sampled skill to maintain balanced skill distribution. Thanks to this selector-based mechanism, multiple skills can be learned simultaneously within a single policy.

\subsection{Training Details}
\label{details}
\subsubsection{Terrain Finetuning}
We design 5 terrain types: flat, slopes, bars, discrete footholds, and stairs, where the latter four are rough terrains. The discrete footholds simulate scenarios where the robot must recover from leg entrapment in gaps. Each type includes 10 difficulty levels, with complexity progressively increased using the terrain curriculum proposed in~\cite{rudin2022learning}. All terrain types are generated before training, with the robot selecting the corresponding terrain during learning. To encourage the robot to lift legs over higher obstacles, we add fractal noise to rough terrains and embed the curriculum with a noise amplitude range of $[0.02, 0.07]$.

\subsubsection{State Initialization}

The design of the initial state plays a crucial role in guiding the robot into more informative regions, accelerating convergence while balancing exploration and exploitation. To leverage keyframes, we randomly initialize the robot’s joint positions with the keyframe $s_i^K$ of the selected skill $i$ (Section.~\ref{skill_initialization}). Notably, the initialized state can differ from the state of the target skill described in Section~\ref{skill_initialization}, which further facilitates seamless skill transitions and mitigates forgetting.

\section{EXPERIMENTS}

The following experiments aim to address:
\begin{itemize}    
    \item Does each component of KiRAS enhance skill diversity or policy robustness? Can KiRAS match the performance of state‑of‑the‑art open‑source algorithms under the same conditions? (In Section.~\ref{problem2})
    \item Can KiRAS learn multiple skills simultaneously, with each achieving ideal performance? (In Section. \ref{problem1})
    \item Is KiRAS capable of traversing complex indoor and outdoor environments with stable and robust performance? How does flexible skill switching benefit locomotion in such environments? (In Section. \ref{problem3})
    \item How well does KiRAS generalize? Can it transfer to quadruped platforms with different DOFs and support a broader range of skill types? (In Section.~\ref{problem4})
\end{itemize}

\subsection{Experimental Setup}
We use Isaac Gym~\cite{makoviychuk2021isaacgymhighperformance} as the simulation platform. In both simulation and real-world experiments, we deploy the Solo-8 robot to perform 5 skills: walking, stilting, crawling, pitching-up, and pitching-down, with their transitions. Each skill has a specific base height and pitch angle summarized in TABLE~\ref{tab:skills}. We train the policy with $T_1 = 6{,}000$ iterations for skill primitives acquisition and $T_2 = 30{,}000$ iterations for robust adaptation, using 4{,}096 parallel environments on an NVIDIA RTX~4090 GPU within 12~hours. After training, we export the policy as an ONNX model (under 500~kB), so it can run on a low-cost onboard computer at 50~Hz.

\begin{table}[h]
\captionsetup{font=small}
\vspace{-0.2cm}
\caption{Parameters of Different Skills.}
\label{tab:skills}
\centering
\begin{tabular}{l|ccccc@{\hspace{-0.2mm}}}
\toprule
\textbf{Skills} & \textbf{Walk} & \textbf{Crawl} & \textbf{Stilt} & \textbf{Pitch-up} & \textbf{Pitch-down} \\
\midrule
\makecell[l]{\textbf{Base Height} (cm)}   & 20 & 10 & 30 & 20 & 20 \\
\makecell[l]{\textbf{Pitch Angle} (deg)}    & 0  & 0  & 0  & -15 & 15 \\
\bottomrule
\end{tabular}
\vspace{-0.4cm}
\end{table}

\subsection{Ablation and Comparison Experiments}
\label{problem2}
\subsubsection{Analysis of Compared Methods}
To comprehensively evaluate the performance of KiRAS, we design 8 comparative baselines in simulation platform:

\begin{itemize}
    \item \textbf{KiRAS w/o SIL}: The policy excludes the Premium Trajectory Selector, manually designing reward functions for each skill based on base height and pitch angle.
    \item \textbf{KiRAS w/o MuC}: The policy excludes MuC and employs a single critic network for value calculation.
    \item \textbf{KiRAS w/o ECE}: The policy excludes ECE while retaining $o_t^H$ as proprioceptive observation input.
    \item \textbf{KiRAS w/o Skill Initialization}: The policy excludes the Skill Initialization Technique and randomly selects a skill to train at the start of each episode.
    \item \textbf{Cassi~\cite{10160421}}: A method that learns skills from unlabeled expert datasets, which collected from open-source algorithms~\cite{li2023learning, 10160421,10160562}.
    \item \textbf{WTW~\cite{margolis2023walk}}: A method encoding a structured family of locomotion policies that solve training tasks in diverse approaches, thereby enabling skill diversity.
    \item \textbf{DreamWaQ~\cite{nahrendra2023dreamwaq}}: A VAE-based method for estimating velocity and future observations, modified with manually designed reward functions to distinguish skills.
    \item \textbf{CaT~\cite{chane2024cat}}: CaT employs an early-terminated DRL algorithm for stable exploration, modified with manually designed constraints to distinguish skills.
\end{itemize}

\begin{figure}[t]
\vspace{0.2cm}
\centerline{\includegraphics[width=\columnwidth]{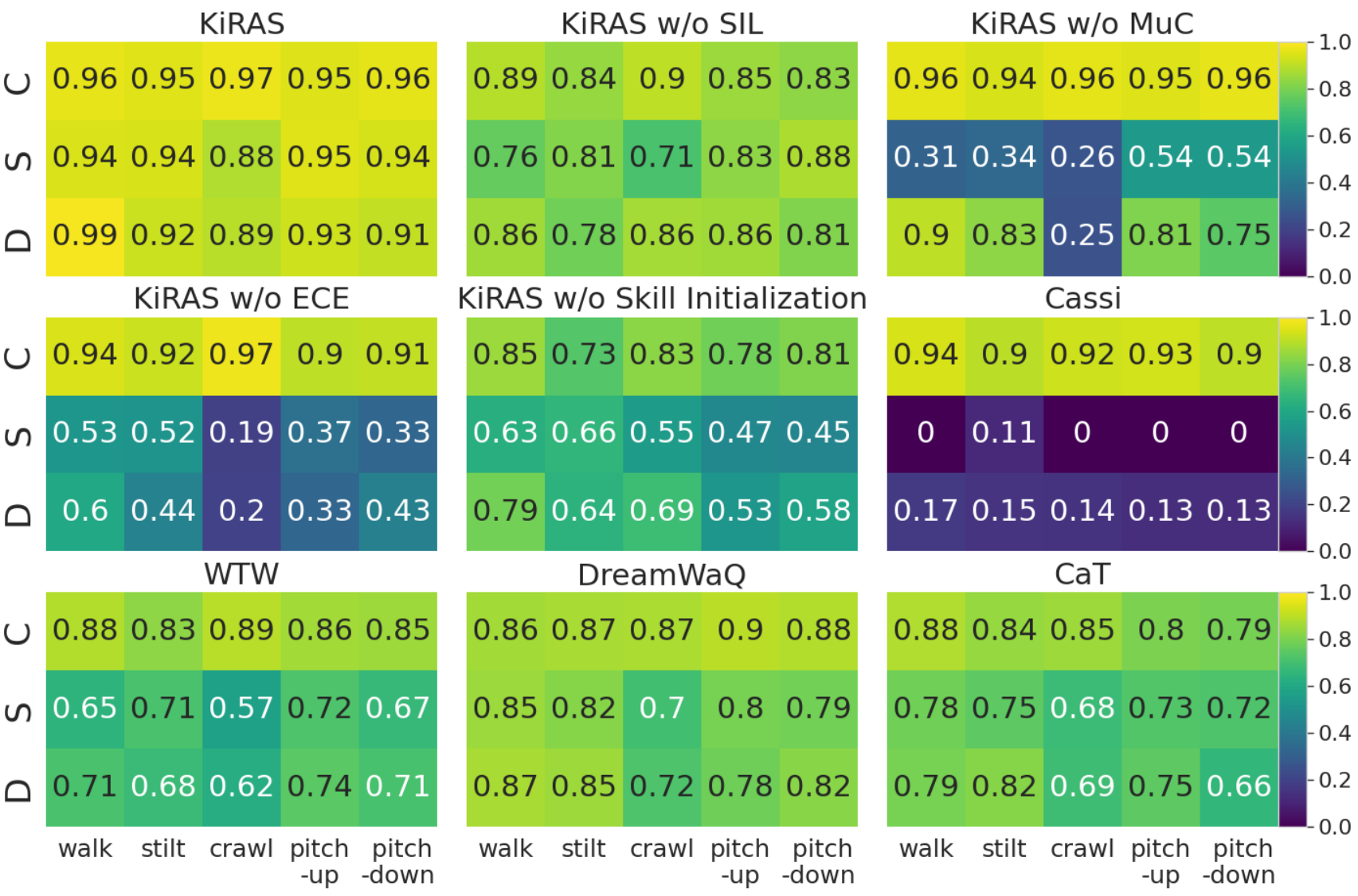}}
\captionsetup{font=footnotesize}
\caption{Heatmap of ablation and comparison experiment results. The vertical axis "C" denotes the cosine similarity between each skill trajectory and its corresponding keyframe, with higher values indicating greater similarity. "S" and "D" represent the success rates of traversing steps and discrete footholds, respectively, where higher values indicate more robust policies.  
}
\vspace{-0.4cm}
\label{heatmap}
\end{figure}

All algorithms are trained for 36{,}000 iterations with identical terrain settings (depicted in Section. \ref{details}) and rewards unless otherwise specified. After training, we compute the cosine similarity (C) between each skill and its corresponding keyframe, which measures the dot-product similarity of trajectories and serves as an indicator of skill acquisition. We further evaluate performance over 1{,}000 trials across arbitrary speeds and orientations, measuring success rates on steps (S) and discrete footholds (D), which quantify terrain adaptability. The resulting heatmap is presented in Fig.~\ref{heatmap}.

\begin{figure}[htbp]
\vspace{-0.2cm}
\centerline{\includegraphics[width=\columnwidth]{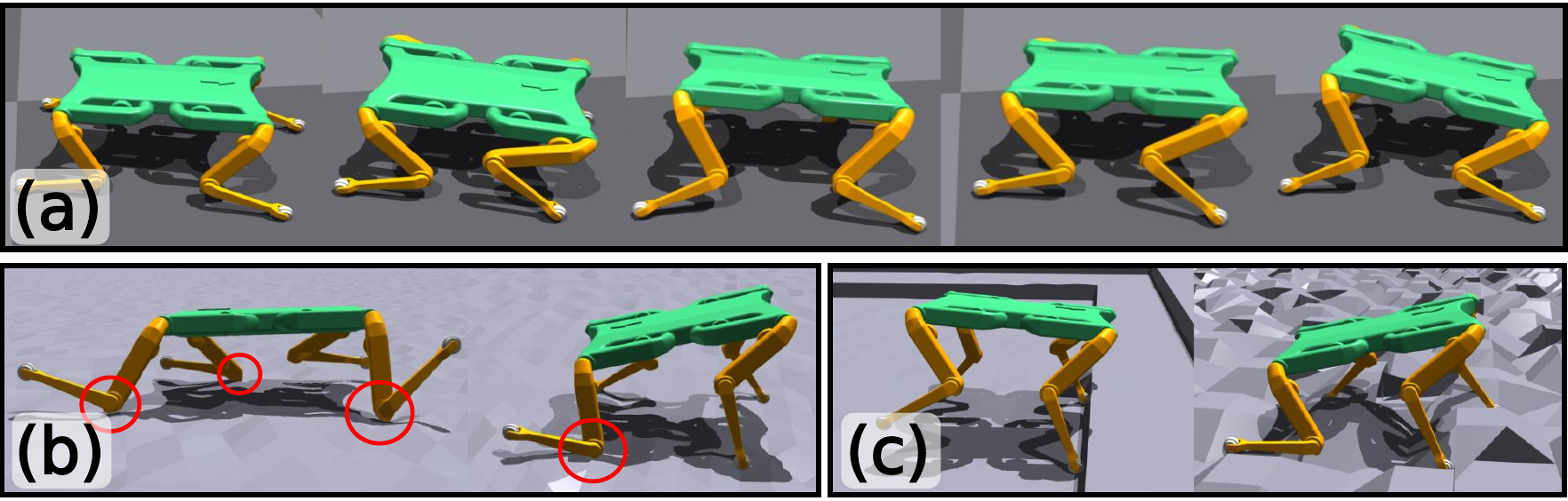}}
\captionsetup{font=footnotesize}
\caption{Comparative results. (a) KiRAS w/o Skill Initialization learns 5 skills (walk, crawl, stilt, pitch-up, pitch-down) with little distinction. (b) DreamWaQ’s crawl and pitch-down use knee-ground contact (red circles). (c) KiRAS w/o MuC fails to lift legs over obstacles.
}
\vspace{-0.2cm}
\label{compared-results}
\end{figure}

With an effective training pipeline and model architecture, KiRAS achieves the highest imitation similarity and superior terrain performance. As shown in the first row of each subfigure, KiRAS maintains dataset-level imitation quality comparable to Cassi, but without relying on large expert datasets. In contrast, Cassi suffers extremely low traversal success due to being overly constrained by the dataset and lacking terrain awareness. This also partially affects Cassi’s skill imitation, leading the policy to converge to unnatural behaviors on challenging terrains. 

\begin{figure*}[htbp]
\vspace{0.2cm}
\centerline{\includegraphics[width=\textwidth]{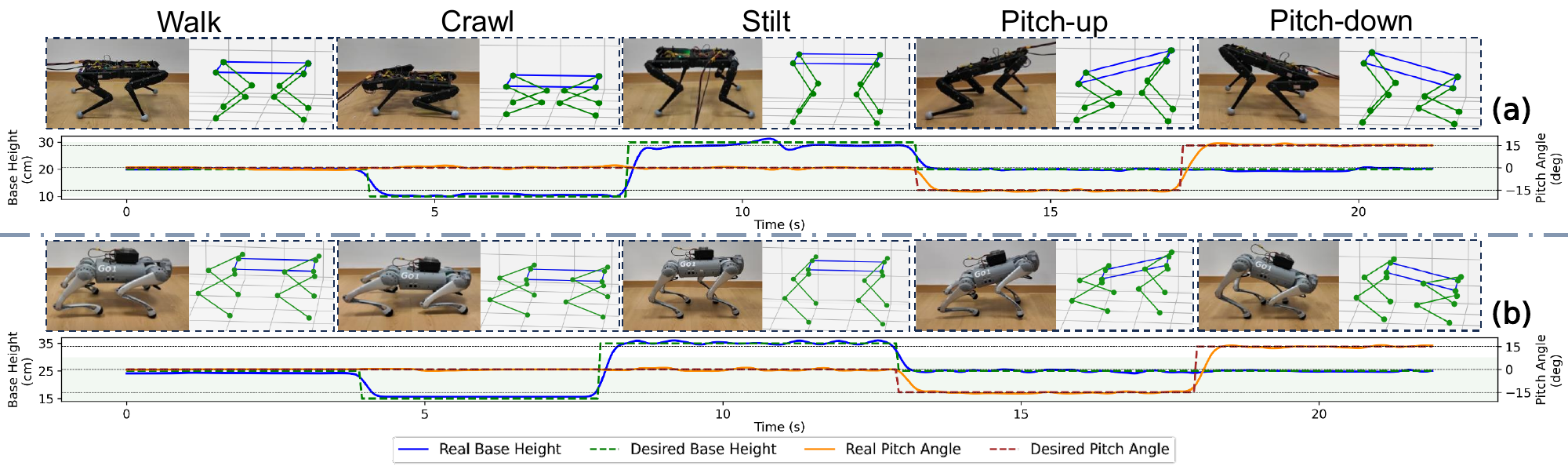}}
\captionsetup{font=footnotesize}
\caption{Real-world experiments validating skill acquisition with KiRAS on the Solo-8 (a) and Unitree Go1 (b) robots. Each black dashed box shows skill execution on hardware, with the left image showing the real robot and the right showing the keyframe used for training. In all experiments, robots are oriented to the right as the forward direction.
}
\vspace{-0.5cm}
\label{imitation-hardware}
\end{figure*}

Removing the Skill Initialization Technique collapses skills into nearly identical gaits that diverge from the keyframes, as illustrated in Fig.~\ref{compared-results}(a), highlighting its role in preventing forgetting. Both KiRAS w/o SIL and DreamWaQ constrain skills via rewards. While effective, they over-constrain the base height or orientation, resulting in the robot contacting the terrain with its knees rather than its feet, especially when the base height is relatively low, as shown in Fig.~\ref{compared-results}(b). This slightly improves success rates but severely compromises hardware safety. Without MuC, the policy remains stuck in static keyframes, failing to learn dynamic foot-lifting and thus struggling on steps, as shown in Fig.~\ref{compared-results}(c). Without ECE, the policy loses proprioceptive terrain awareness and performs like WTW; although WTW distinguishes skills, neither approach has terrain perception and both fail on challenging terrains. CaT relies on constrained terminations but is not designed for traversal, limiting its performance.

\subsubsection{Effectiveness of Skill Initialization Technique}

\begin{figure}[t]
\vspace{0.2cm}
\centerline{\includegraphics[width=\columnwidth]{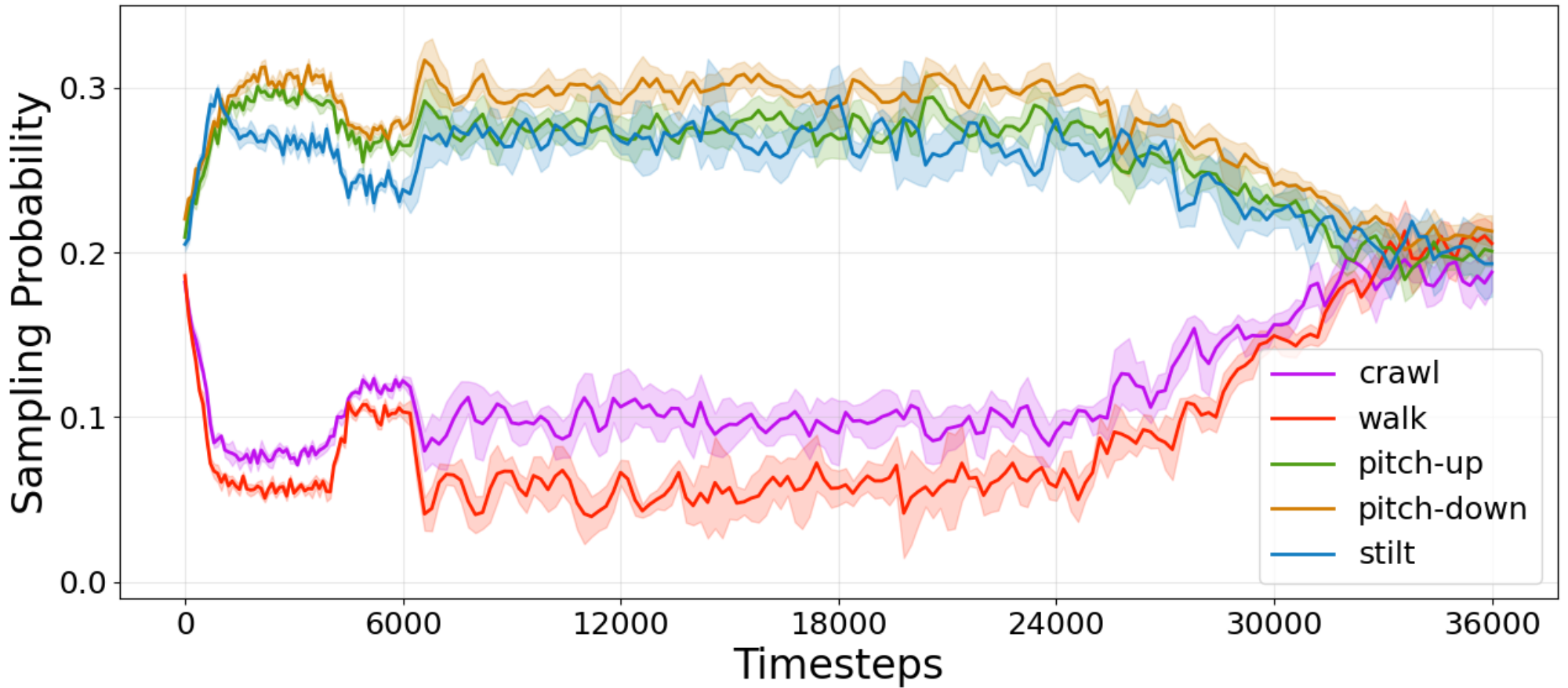}}
\captionsetup{font=footnotesize}
\caption{Sampling probability evolution for each skill during training, computed using 4 random seeds. Curves and shaded areas represent the mean and standard deviation across seeds, respectively.}
\vspace{-0.6cm}
\label{selector}
\end{figure}

This experiment evaluates the Skill Initialization Technique for balancing multi-skill learning. Fig.~\ref{selector} shows the evolution of sampling probability $p(\text{Skill}=i)$ for 5 skills over 36{,}000 iterations. During skill learning, all skills converge at about 4,500 iterations with probabilities near 0.2. During terrain finetuning, variations demand relearning adaptability. Among these skills, walk and crawl converge faster due to lower center of mass, reduced motion range, and balance demands, leading to lower probabilities. In contrast, stilt requires higher center of mass and extended limbs, demanding precise control to avoid instability and thus receiving higher probability. Pitch-up and pitch-down require stable pitch and altered force distribution, increasing correction difficulty and yielding the highest probabilities. The Solo-8’s front-heavy design makes pitch-down more prone to slipping or overload, raising its probability. To prevent forgetting, all skills maintain non-zero sampling throughout training. These results show KiRAS adaptively adjusts sampling to skill difficulty, ensuring balanced multi-skill learning.

\subsection{Skill Imitation Sim-to-real Experiments}
\label{problem1}

To evaluate skill acquisition, we extract the policy and deploy it on the real Solo-8 robot, comparing the skills with the corresponding keyframes. Moreover, we record the base height and pitch angle while the robot executes skills and transitions. As shown in Fig.~\ref{imitation-hardware}(a), the executed motions closely match the designed keyframes, confirming the accuracy in reproducing the intended motion style. The tracked values further align with TABLE~\ref{tab:skills}, verifying precise behaviors under keyframe constraints. In addition, during transitions between different skills, the trajectories remain smooth and continuous, without noticeable oscillations or delays, demonstrating remarkable stability and controllability in dynamic switching.

Overall, these results show that KiRAS efficiently learns and transitions among multiple skills from keyframe guidance alone. The comparable similarity to large-scale datasets highlights the potential as a compact, keyframe-driven platform for generating multi-skill quadruped demonstrations, enabling efficient collection of complex locomotion tasks.

\subsection{Indoor and Outdoor Sim-to-real Experiments}
\label{problem3}

\begin{figure*}[htbp]
\vspace{0.2cm}
\centerline{\includegraphics[width=\textwidth]{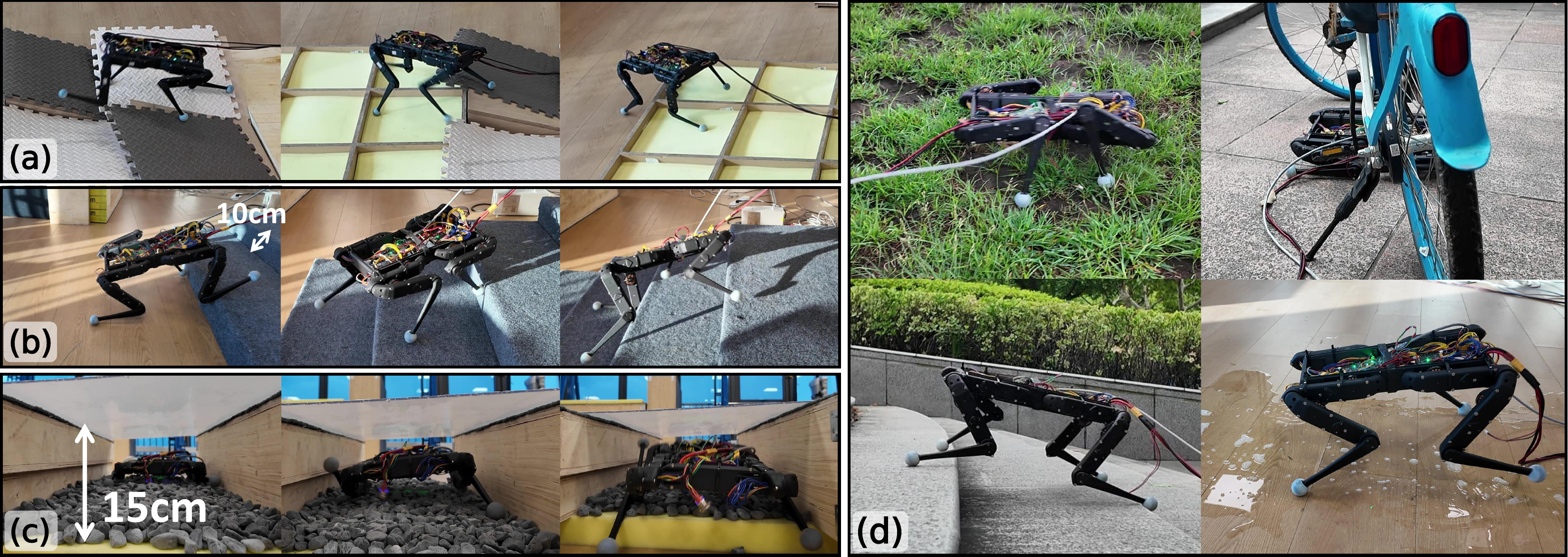}}
\captionsetup{font=footnotesize}
\caption{Solo-8 robot traversing diverse and challenging terrains in indoor and outdoor environments with different skills. }
\vspace{-0.6cm}
\label{solo8-indoor}
\end{figure*}

Extensive indoor and outdoor experiments demonstrate KiRAS’s effectiveness, robustness, and versatility in sim-to-real transfer, enabling traversal of diverse terrains with multiple skills. These results highlight the importance of flexible skill switching for locomotion in complex environments. To our knowledge, this is the first work to achieve agile locomotion control for an 8-DoF quadruped robot on complex terrain.

\subsubsection{Locomotion Ability Test}

Fig.~\ref{solo8-indoor} shows the Solo-8 performing multi-skill locomotion: (a) walking across wedge-shaped obstacles and soft ground; (b) climbing a 10\,cm step with pitch-up; (c) crawling through a 15\,cm-high stone-filled passage; and (d) crawling through pit-filled grass and a bicycle obstacle, while descending steps and traversing slippery surfaces by the walk skill. All tasks require lifting the legs to overcome elevated obstacles, which emerges in the terrain finetuning even though it is not explicitly encoded in the keyframes. Solo-8 is compact (2\,kg), with limited joint torque (1\,Nm, about $1/30$ of Unitree Go1) and no hip joints, making rough-terrain locomotion challenging. However, KiRAS enables it to complete demanding tasks (Fig.~\ref{solo8-indoor}(a,b)). The lack of joint limits further allows extremely low postures, enabling traversal through narrow gaps (Fig.~\ref{solo8-indoor}(c,d-upper); Fig.~\ref{performance}(c)). Moreover, its lightweight design provides stability on slippery surfaces through improved center-of-mass control (Fig.~\ref{solo8-indoor}(d-lower)). These results demonstrate the robustness of the policies trained with KiRAS, enabling the robot to traverse complex terrains with diverse skills.

\subsubsection{Terrain Adaptation of Individual Skills}

To verify that specific skills better adapt to challenging terrains, we conduct 50 experiments on a real Solo-8 robot, using different skills to climb 45° slopes and traverse 15 cm steps. These tasks are highly challenging for the Solo-8 robot. The success rates, summarized in TABLE~\ref{tab:terrain}, show that pitch-down achieves the highest rate on steep slopes by lowering the front body to reduce tipping and extending the hind legs for stronger propulsion. In contrast, pitch-up excels on stairs, as lifting the forelegs facilitates step clearance while stabilizing the center of mass. Other skills exhibit limitations in either stability or obstacle clearance, resulting in lower success rates.  

\begin{table}[h]
\captionsetup{font=small}
\vspace{-0.2 cm}
\caption{Success rates of Solo-8 robot traversing terrains with different skills. The highest rates are in bold.
}
\label{tab:terrain}
\centering
\begin{tabular}{l|ccccc@{\hspace{-0.2mm}}}
\toprule
\textbf{Skills} & \textbf{Walk} & \textbf{Crawl} & \textbf{Stilt} & \textbf{Pitch-up} & \textbf{Pitch-down} \\
\midrule
\makecell[l]{\textbf{Slope}}   & 0.84 & 0.68 & 0.80 & 0.70 & \textbf{0.9} \\
\makecell[l]{\textbf{Upstairs}}    & 0.80  & 0.54  & 0.44  & \textbf{0.88} & 0.58 \\
\makecell[l]{\textbf{Downstairs}}    & 0.88  & 0.46  & 0.70  & \textbf{0.94} & 0.56 \\
\bottomrule
\end{tabular}
\vspace{-0.2cm}
\end{table}

\begin{figure}[htbp]
\vspace{0.2cm}
\centerline{\includegraphics[width=\columnwidth]{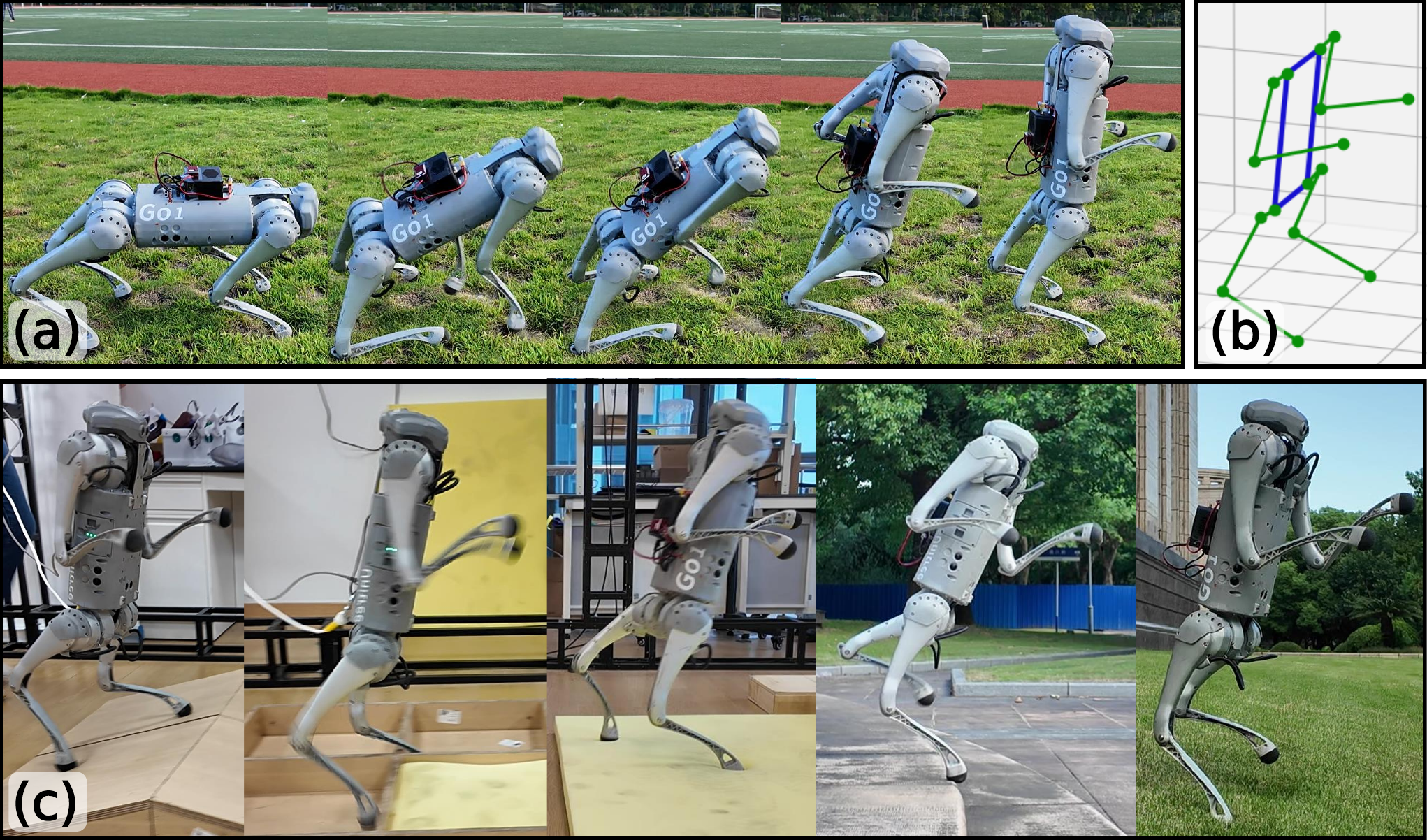}}
\captionsetup{font=footnotesize}
\caption{(a) Transition from the walk skill to the biped skill. (b) Keyframe for the biped skill used for imitation. (c) Snapshots of executing the biped skill to traverse various indoor and outdoor terrains. 
}
\vspace{-0.5cm}
\label{biped}
\end{figure}

To further illustrate this, Fig.~\ref{performance}(a) shows the robot employing different skills: crawl for a low bar, stilt for a tall barrier, pitch-down for ascending a slope, and pitch-up for descending stairs. The first two tasks are posture-constrained, with crawl and stilt required by geometry, demonstrating the role of base height in restricted settings. The latter highlight flexibility, consistent with quantitative results above. These results confirm that skills provide task-specific solutions to improve stability across diverse terrains.

\subsection{Generalization Experiments}
\label{problem4}

\subsubsection{Generalization to Robot with Different DOFs}

Since KiRAS relies solely on single-frame keyframes, its reference data is easily accessible, enabling cross-platform skill training. We train a policy on Unitree Go1 robot by adjusting joint positions of Solo-8 keyframes, achieving accurate multi-skill reproduction (Fig.~\ref{imitation-hardware}(b)). Because Unitree Go1 is larger than Solo-8, all keyframes are designed with a 5\,cm higher base. Similar to Section~\ref{problem1}, we record base height and pitch angle during continuous trajectories. Results show that KiRAS can preserve precise skill execution and smooth transitions on a robot with different DOFs.

\subsubsection{Generalization to New Skills by New Keyframes}
To evaluate the generalization of KiRAS, we train an additional skill that can be represented by one keyframe: biped (Fig.~\ref{biped}(b)). This task requires the robot to stand upright on its hind legs, which is more challenging due to reduced stability, requiring continuous gait adjustments for balance. In our experiments, we extend the existing policy trained on 5 skills by incorporating the keyframes of the new skill, followed by training for $T_1=10{,}000$ and $T_2=30{,}000$ timesteps to progressively integrate it into the original framework.

As shown in Fig.~\ref{biped}(a), when transitioning from walking to bipedal stance, the robot first raises its front body and shifts the center of mass to the hind legs. It then pushes off with its forelegs to gain upward momentum, bringing the base close to vertical. At this stage, the robot relies on continuous hind-leg adjustments to maintain balance. This demonstrates that KiRAS can effectively learn more challenging skills while maintaining stable control.

Furthermore, we evaluate the robot’s biped skill on a variety of unstructured terrains, including smooth slopes, discrete stepping stones, sponges, stairs, sloped grassy surfaces, and playgrounds with pits and gravel, as illustrated in Fig.~\ref{performance}(d), and Fig.~\ref{biped}(c). Among these, stairs and discrete terrains have proven to be the most challenging. Nevertheless, thanks to the robustness of the policy, the bipedal robot can safely traverse stairs up to 10\,cm in height. These results further demonstrate that KiRAS enables stable locomotion in complex environments.

\section{CONCLUSIONS}

This work presents KiRAS, a keyframe-driven end-to-end DRL framework that unifies multi-skill learning and terrain adaptability. KiRAS acquires diverse skills from single keyframes within a unified policy,  and progressively masters locomotion tasks through skill primitives. The Skill Initialization Technique further mitigates catastrophic forgetting by efficiently selecting subsequent training actions. Experiments show that KiRAS achieves strong skill diversity and terrain adaptability, matching open-source methods built on large datasets. It also generalizes to quadruped platforms with varying DOFs and new skill types, highlighting its potential as a lightweight platform for multi-skill generation and dataset collection. Future work will explore adaptive skill switching using environmental cues, and extend the framework to navigation tasks with exteroceptive sensing.




\bibliographystyle{IEEEtran}
\bibliography{mylib}

\section{APPENDIX}
\subsection{Observation and Reward Settings}
\subsubsection{Observation}
We adopt an asymmetric actor-critic architecture with self-imitation signals. 
The policy receives proprioceptive observations and a latent context inferred from short-term history. 
At timestep $t$, the actor input is $ o_t^{\pi} = [\,o_t^{\mathrm{prop}},\, z_t\,], $where $o_t^{\mathrm{prop}}$ denotes the current proprioceptive observation.  The latent context $z_t \in \mathbb{R}^{11}$ is encoded from the past 4 proprioceptive observations 
$\{o_{t-4}^{\mathrm{prop}}, \dots, o_{t-1}^{\mathrm{prop}}\}$ through a VAE-style encoder.
Specifically, the latent vector is defined as $ z_t = [\,\hat{v}_t,\, z_t^{\mathrm{lat}}\,], $
where $\hat{v}_t \in \mathbb{R}^{3}$ is a supervised estimate of the base linear velocity and 
$z_t^{\mathrm{lat}} \in \mathbb{R}^{8}$ is a stochastic latent variable capturing residual motion context. The critic receives privileged observations $o_t^{V}$ that augment proprioceptive signals with additional information useful for value estimation (e.g., terrain height samples). For imitation learning and discriminator training, we construct a compact imitation observation $\Phi(o_t)$ composed of root height, joint states, and skill indicators. The detailed observation components are summarized in Table~\ref{tab:obs_terms}.

\begin{table}[h]
\captionsetup{font=small}
\caption{Observation components.}
\label{tab:obs_terms}
\centering
\begin{tabular}{lccc}
\toprule
\textbf{Observation Term} & $\mathbf{o_t^{\mathrm{prop}}}$ & $\mathbf{o_t^{V}}$ & $\mathbf{\Phi(o_t)}$ \\
\midrule
Base linear velocity &  & \checkmark &  \\
Base angular velocity &  & \checkmark &  \\
Foot height &  & \checkmark &  \\
Projected gravity &  & \checkmark &  \\
Base orientation & \checkmark & \checkmark & \checkmark \\
Command & \checkmark & \checkmark &  \\
Skill vector & \checkmark & \checkmark & \checkmark \\
Joint position &  &  & \checkmark \\
Joint position error & \checkmark & \checkmark &  \\
Joint velocity & \checkmark & \checkmark &  \\
Previous action & \checkmark & \checkmark &  \\
Terrain height samples &  & \checkmark &  \\
Base height &  &  & \checkmark \\
\midrule
History encoder input & \checkmark &  &  \\
Latent context $z_t$ (actor input) & \checkmark &  &  \\
\bottomrule
\end{tabular}
\end{table}

\subsubsection{Reward}
In KiRAS, the total reward consists of four components: the self-imitation learning (SIL) reward $r_t^{{SI}}$ for learning skill primitives, the termination penalty $r_t^T$, the regularization reward $r_t^R$ for stable real-world deployment, and a residual penalty $r_t^{{res}}$ that constrains joint outputs. 

To compute the SIL reward, we train a discriminator $D_\psi$ to distinguish state transitions $(\Phi(o_{t-1}), \Phi(o_t))$ sampled either from the premium buffer distribution $B_g$ or from the policy distribution $d_\pi$:
\begin{equation}
\begin{aligned}
    \min_{\psi} \; &\mathbb{E}_{(\Phi(o_{t-1}), \Phi(o_t)) \sim B_g} 
    \left[ \left(D_\psi(\cdot) - 1\right)^2 \right] \\
    +\; &\mathbb{E}_{(\Phi(o_{t-1}), \Phi(o_t)) \sim d_\pi} 
    \left[ \left(D_\psi(\cdot) + 1\right)^2 \right].
\end{aligned}
\end{equation}

Unlike WASABI, we do not apply gradient smoothing to the discriminator. Since the discriminator only needs to distinguish among a small set of skills, the classification problem remains relatively simple, and additional smoothing may unnecessarily slow down training without providing substantial benefit. The SIL reward is then defined as
\begin{equation}
   r_t^{SI} = \max \left[1 - 0.25\left(D_\psi(\Phi(o_{t-1}), \Phi(o_t)) - 1\right)^2, \; 0\right],
\end{equation}
which provides a reward signal encouraging the policy to generate transitions that resemble high-return trajectories in the premium buffer, thereby guiding self-imitation and facilitating effective skill learning.

The detailed formulations and coefficients of the remaining reward components are summarized in Table~\ref{tab:rewards}.
\begin{table}[h]
\captionsetup{font=small}
\caption{Reward terms and coefficients.}
\label{tab:rewards}
\centering
\setlength{\tabcolsep}{20pt}
\begin{tabular}{lc}
\toprule
\textbf{Reward name} & \textbf{Coefficient} \\
\midrule

\multicolumn{2}{c}{\textbf{Termination penalty} $r_t^T$} \\
\midrule
collision & -0.5 \\

\midrule
\multicolumn{2}{c}{\textbf{Regularization reward} $r_t^R$} \\
\midrule
feet\_contact\_forces & -1.0 \\
action\_rate & -0.2 \\
torques & $-2.5\times10^{-5}$ \\
delta\_torques & $-1.0\times10^{-3}$ \\
tracking\_lin\_vel & 1.0 \\
tracking\_ang\_vel & 0.5 \\
feet\_drag & 0.5 \\
ang\_vel & -0.1 \\
stand\_still & -0.2 \\
joint acceleration & $-1.25\times10^{-8}$ \\
angular velocity x & -0.001 \\
angular velocity z & -0.001 \\
linear velocity y & -0.001 \\

\midrule
\multicolumn{2}{c}{\textbf{Residual reward}} \\
\midrule
$r_t^{\mathrm{res}}$ & 0.15 \\

\bottomrule
\end{tabular}
\end{table}

\subsection{More Experiment Details}
\subsubsection{Domain Randomization}
To improve robustness in rough-terrain locomotion, we randomize robot and environment properties during training. 
Specifically, we randomize contact friction, base mass, base center-of-mass (CoM), external pushes, and actuation delay. 
Friction is sampled through bucketed randomization across environments; base mass and CoM are perturbed at reset; and random pushes are periodically injected by perturbing base velocity. 
In addition, we model control-side timing uncertainty via random action delay within one control decimation window (4 simulation steps). 
Given the simulator step size $\Delta t = 5$ ms, this corresponds to a delay range of $[0,15]$ ms. 
The exact ranges used in our implementation are listed in Table~\ref{tab:dr_ranges}.

\begin{table}[h]
\captionsetup{font=small}
\caption{Domain randomization settings.}
\label{tab:dr_ranges}
\centering
\setlength{\tabcolsep}{15pt}
\begin{tabular}{lc}
\toprule
\textbf{Parameter} & \textbf{Range / Value} \\
\midrule
Base payload perturbation & $[-0.5,\,0.25]$ kg \\
Base CoM shift (per axis) & $[-0.02,\,0.05]$ m \\
Contact friction coefficient & $[0.5,\,1.25]$ \\
External push velocity (base) & $[-0.1,\,0.1]$ m/s \\
Push interval & $5$ s \\
Actuation delay & $[0,\,15]$ ms \\
PD stiffness multiplier & $[0.8,\,1.2]$ \\
PD damping multiplier & $[0.8,\,1.2]$ \\
\bottomrule
\end{tabular}
\end{table}

\subsubsection{Network Parameters}
Our policy adopts an asymmetric actor-critic architecture with two critics and ECE, together with an imitation discriminator. 
The actor receives proprioceptive observations augmented with latent context inferred from short-term history, while the critics use privileged observations for value estimation. The actor-critic and discriminator are jointly optimized using Adam ($\text{lr}=10^{-3}$). 
Detailed network architectures used for training the Solo-8 policy are summarized in Table~\ref{tab:net_arch}.

\begin{table}[h]
\captionsetup{font=small}
\caption{Network architectures for training Solo-8 policy.}
\label{tab:net_arch}
\centering
\begin{tabular}{lcc}
\toprule
\textbf{Network} & \textbf{In/Out} & \textbf{Hidden} \\
\midrule
Actor & $42 \rightarrow 8$ & [128,128,128] \\
Task critic $V_{\phi_1}$ & $107 \rightarrow 1$ & [128,128,128] \\
Optimization critic $V_{\phi_2}$ & $107 \rightarrow 1$ & [128,128,128] \\
ECE encoder & $124 \rightarrow 19$ & [128,64] \\
ECE decoder & $13 \rightarrow 31$ & [64,128] \\
Imitation discriminator & $38 \rightarrow 1$ & [512,256] \\
\bottomrule
\end{tabular}
\end{table}

\subsection{More Results}
\subsubsection{Quantitative Experiments on Skill Imitation}

To quantify skill acquisition performance, we extract the learned policy of the Solo-8 robot at $T_1$ iterations from the simulation and conduct a comprehensive evaluation using 100 parallel robot instances, each executing a trajectory of 120 timesteps.  We compute two evaluation metrics: (1) DTW with $L_2$ norm, also employed for premium trajectory selection, and (2) Cosine Similarity, capturing dot-product similarity. Learned trajectories are compared with two references: (i) \textit{Keyframe}, the initial training keyframes, and (ii) \textit{Dataset}, diverse demonstrations from~\cite{li2023learning} and~\cite{10160421}. As presented in Table \ref{tab:dtw_cosine}, KiRAS successfully reproduces all keyframe-guided skills and achieves DTW and Cosine Similarity scores that are comparable to, or even outperform, those of the expert demonstration dataset. This result further validates that KiRAS enables accurate multi-skill learning with minimal supervisory signals.

\begin{table}[h]
\captionsetup{font=small}
\caption{Comparison of DTW (lower is better) and Cosine Similarity (higher is better) across skills.}
\label{tab:dtw_cosine}
\centering
\begin{tabular}{l|cccccc}
\toprule
 \textbf{Skills}  & \textbf{Walk} & \textbf{Stilt} & \textbf{Crawl} & \textbf{Pitch-up} & \textbf{Pitch-down} \\
\midrule
\multicolumn{6}{l}{\textbf{DTW} $\downarrow$} \\
Keyframe   & \textbf{15.67} & \textbf{15.19} & \textbf{20.11} & \textbf{18.03} & \textbf{19.04} \\
Dataset    & 16.42          & 15.84             & 24.18          & 19.53          & 19.96 \\
\midrule
\multicolumn{6}{l}{\textbf{Cosine Similarity} $\uparrow$} \\
Keyframe   & \textbf{0.978} & \textbf{0.974} & \textbf{0.987} & \textbf{0.965} & \textbf{0.976} \\
Dataset    & 0.963          & 0.942             & 0.966          & 0.949          & 0.953 \\
\bottomrule
\end{tabular}
\vspace{-0.2cm}
\end{table} 

\subsubsection{Sim-to-real Experiments on Skill Transition}
We further evaluate the sim-to-real performance of KiRAS on real hardware by demonstrating skill transitions using a single all-in-one policy. 
As shown in the accompanying videos on the project website, the robot can smoothly switch among different skills without resetting or loading separate controllers. 
The transitions are triggered online through the skill command while maintaining stable locomotion, illustrating that the learned policy captures coherent behaviors across multiple skills. 
These results demonstrate that KiRAS enables reliable and fluid arbitrary skill transitions under a unified policy in the real world.


\end{document}